\documentclass[10pt,twocolumn,letterpaper]{article}

\usepackage{wacv}              % To produce the CAMERA-READY version

\usepackage{xcolor}
\usepackage{pifont} % Required for \ding

\newcommand{\cmark}{\textcolor{green}{\ding{51}}}
\newcommand{\xmark}{\textcolor{red}{\ding{55}}}

\usepackage{pifont}
\providecommand{\cmark}{\ding{51}}%
\providecommand{\xmark}{\ding{55}}%

\definecolor{wacvblue}{rgb}{0.21,0.49,0.74}
\usepackage[pagebackref,breaklinks,colorlinks,allcolors=wacvblue]{hyperref}

\def\wacvPaperID{5} % *** Enter the WACV Paper ID here
\def\confName{WACV}
\def\confYear{2026}

\relax\title{\textit{eSkiTB}: A Synthetic Event-based Dataset for Tracking Skiers}

\author{Krishna Vinod \quad Joseph Raj Vishal \quad Kaustav Chanda\\
Prithvi Jai Ramesh \quad Yezhou Yang \quad Bharatesh Chakravarthi\\
Arizona State University\\
{\tt\small \{kvinod, jnolas77, kchanda3, pjramesh, yz.yang, bshettah\}@asu.edu}
}

\usepackage{background}

\newcommand{\mywatermark}{%
    \begin{minipage}{\textwidth}
        \centering
        \fontsize{10}{10}\selectfont % Adjust font size and baselineskip
        This paper has been accepted at the WACV 2026 Workshop on Computer Vision for Winter Sports
    \end{minipage}%
}

\backgroundsetup{
    scale=1,
    color=gray,
    angle=0,
    opacity=0.5,
    position=current page.north,
    vshift=-1cm, % Adjust vertical position
    contents={\mywatermark}
}

\begin{document}
\twocolumn[{%
\renewcommand\twocolumn[1][]{#1}%
\maketitle

\begin{center}
    \centering
    \captionsetup{type=figure}
    \includegraphics[width=0.85\linewidth]{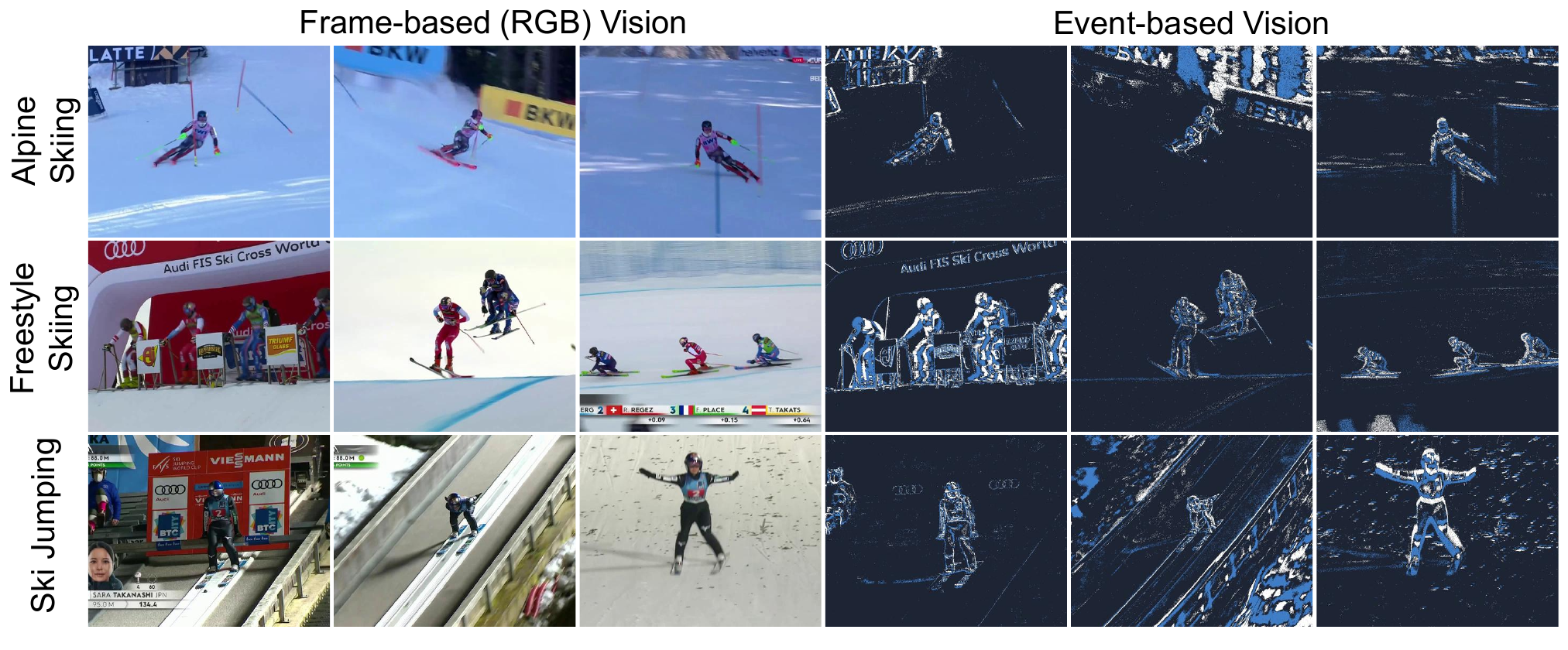}
    \captionof{figure}{Represents a visual overview of the \textit{eSkiTB} dataset. The RGB frame illustrates the challenge of background clutter and broadcast artifacts, while the corresponding event stream demonstrates the natural filtering of static distractions like text overlays and fences.}
    \label{fig:teaser}
\end{center}%
}]

\begin{abstract}

Tracking skiers in RGB broadcast footage is challenging due to motion blur, static overlays, and clutter that obscure the fast-moving athlete. Event cameras, with their asynchronous contrast sensing, offer natural robustness to such artifacts, yet a controlled benchmark for winter-sport tracking has been missing. We introduce event SkiTB (eSkiTB), a synthetic event-based ski tracking dataset generated from SkiTB using direct video-to-event conversion without neural interpolation, enabling an iso-informational comparison between RGB and event modalities. Benchmarking SDTrack (spiking transformer) against STARK (RGB transformer), we find that event-based tracking is substantially resilient to broadcast clutter in scenes dominated by static overlays, achieving $0.685$ IoU, outperforming RGB by $+20.0$ points. Across the dataset, SDTrack attains a mean IoU of $0.711$, demonstrating that temporal contrast is a reliable cue for tracking ballistic motion in visually congested environments. eSkiTB establishes the first controlled setting for event-based tracking in winter sports and highlights the promise of event cameras for ski tracking. The dataset and code will be released at \url{https://github.com/eventbasedvision/eSkiTB}.

\end{abstract}

\section{Introduction}
\label{sec:intro}

\begin{table*}[ht]
\centering
\caption{Summarises \textit{eSkiTB} against benchmarks, highlighting its unique non-rigid motion, and cluttered broadcast environments.}
\resizebox{0.94\linewidth}{!}{%
\begin{tabular}{@{}lccccccc@{}}
\toprule
\textbf{Dataset} & \textbf{Sensor} & \textbf{Resolution} & \textbf{Target Type} & \textbf{Motion Dynamics} & \textbf{Background} & \textbf{Key Challenge} & \textbf{Suitability for Sports} \\ \midrule
\multicolumn{8}{c}{\textbf{RGB (frame-based) Benchmarks}} \\
OTB-100~\cite{wu2015object} & RGB & $640 \times 480$ & Rigid / Mixed & Linear/Planar & Cluttered & Occlusion & Low  \\
LaSOT~\cite{fan2019lasot} & RGB & Various & Mixed & Standard & General & Long-term & Medium \\ \midrule
\multicolumn{8}{c}{\textbf{Event-based Benchmarks}} \\
FE108~\cite{zhang2021fe108} & DVS128 & $128 \times 128$ & Rigid/Mixed & High Speed & Lab/Outdoor & Temporal Res & Low  \\
VisEvent~\cite{wang2021visevent} & DVS346 & $346 \times 260$ & Mixed & Standard & General & Low Light & Medium \\
COESOT~\cite{tang2022coesot} & DVS346 & $346 \times 260$ & Mixed & Standard & General & Scale & Medium \\ \midrule
\textbf{\textit{eSkiTB} (Ours)} & \textbf{\textit{v2e} (Sim)} & \textbf{$1280 \times 720$} & \textbf{Non-Rigid Human} & \textbf{Ballistic ($>$$\textbf{100}$km/h)} & \textbf{Cluttered Broadcast} & \textbf{Broadcast Overlays} & \textbf{High} \\ \bottomrule
\end{tabular}%
}
\label{tab:dataset_comparison}
\end{table*}

Professional winter sports present a uniquely challenging domain for computer vision. The environment is characterized by extreme dynamics, where athletes in disciplines such as ski jumping achieve ballistic velocities exceeding $100$\, km/h~\cite{Bachmann2019SkiPose}. This challenge is compounded by \textit{broadcast clutter} such as static digital overlays, scorecards, and sponsorship banners that are ubiquitous in public datasets sourced from television broadcasts \cite{vstepec2022video, skitb_wacv2024}. In such scenarios, traditional RGB tracking algorithms, which rely heavily on consistent texture and color histograms, frequently suffer catastrophic drift due to \textit{distractor confusion} caused by these persistent visual artifacts. 

Event cameras, or neuromorphic sensors \cite{Gallego2020Survey, chakravarthi2024recent}, offer a robust alternative by asynchronously recording pixel-level brightness changes. This sensing principle introduces a key physical advantage for broadcast-based winter sports. Digital overlays such as scoreboards and graphics remain fixed in the image plane and therefore produce no events, effectively disappearing from the stream as shown in \autoref{fig:teaser}. 
In contrast, physical clutter like trackside banners generates events only due to camera panning, and thus follows the global ego-motion of the background. The athlete, however, produces independent motion, creating a distinct event signature that allows event-based trackers to reliably separate the target from background flow. Despite these promising characteristics, event cameras have not been used in professional winter sports, and no labeled datasets exist for this domain. This lack of data has prevented the community from assessing whether spiking neural networks (SNNs) truly outperform conventional convolutional neural networks (CNNs) based trackers in low-texture, cluttered environments. Crucially, there is still no controlled benchmark that enables a fair, information-equivalent comparison between RGB and event modalities.

To address this gap, we introduce \textit{event SkiTB} (\textit{eSkiTB}), a benchmark designed to enable systematic evaluation of event-based tracking in winter sports. We generate synthetic event streams from the \textit{SkiTB} dataset~\cite{Bachmann2019SkiPose} under a strict \textit{iso-informational constraint} using \textit{v2e} event-simulator~\cite{hu2021v2e}. 
\textit{eSkiTB} provides high-resolution ($1280\times720$) event streams generated directly from Olympic ski broadcast footage. The dataset captures the full visual complexity of this domain, including extreme ballistic velocities, aggressive camera panning, dense broadcast overlays, and diverse weather conditions, while also spanning multiple camera viewpoints that introduce significant scale variation and occlusions. By preserving the raw temporal dynamics of the original footage, \textit{eSkiTB} serves as a testbed for evaluating tracker robustness against the unique visual clutter present in professional winter sports broadcasts.

Unlike prior approaches that rely on neural frame interpolation to create additional temporal cues, our conversion pipeline produces events directly from raw frames, without any intermediate synthesis ~\cite{hu2021v2e, gehrig2018esim}. This ensures that the event stream contains no more temporal information than the original RGB footage, enabling a fair and rigorous comparison between sensing modalities.
Using this benchmark, we evaluate two \textit{state-of-the-art} (SOTA) contrasting tracking paradigms, \textit{STARK}~\cite{yan2021learning}, a RGB transformer that relies on appearance cues, and \textit{SDTrack}~\cite{shan2025sdtrack}, a spiking transformer that exploits the temporal contrast inherent to event data. We summarize our contributions as follows.
\begin{enumerate}
    \item \textbf{\textit{eSkiTB dataset:}} We offer the first large-scale synthetic event-based dataset for ski jumper tracking~\cite{DUNNHOFER2024103978}, generated via a high-fidelity simulation pipeline~\cite{hu2021v2e}.
    \item \textbf{\textit{Iso-informational conversion:}} We propose a controlled dataset-generation protocol that removes neural interpolation, ensuring that any performance gains arise solely from the event modality’s temporal contrast representation.
    \item \textbf{\textit{Event vs. RGB benchmarking:}} We provide a systematic comparison of RGB and event-based trackers in broadcast-heavy winter sports, demonstrating that event-based tracking is more robust to static overlays and other visual distractors.
\end{enumerate}

The remainder of this paper is organized as follows. Section~\ref{sec:background} reviews event-based vision for skiing and summarizes the relevant tracking paradigms. Section~\ref{sec:dataset} details the construction of \textit{eSkiTB}, including the event-generation pipeline, dataset composition, annotations, statistics, and dense temporal ground truth. Section~\ref{sec:evaluation} describes the experimental setup and presents benchmarking results comparing RGB and event-based trackers. Section~\ref{sec:discussion} discusses limitations and broader implications, and Section~\ref{sec:conclusion} summarizes key findings and outlines future directions.

\begin{table*}[t!]
\centering
\caption{Summarises synthetic event datasets generated from RGB videos. \textit{eSkiTB} uniquely provides iso-informational conversion and focuses on tracking in challenging winter environments.}
\resizebox{0.88\linewidth}{!}{%
\begin{tabular}{@{}lcccccc@{}}
\toprule
\textbf{Dataset} & \textbf{Source (RGB)} & \textbf{Simulator} & \textbf{Task} & \textbf{Resolution} & \textbf{Iso-Informational} & \textbf{Harsh Env.} \\ \midrule
Event-KITTI~\cite{gehrig2018esim} & KITTI~\cite{geiger2013vision} & ESIM & Odometry/SLAM & $1241 \times 376$ & \xmark & \xmark \\
Mono-Dataset~\cite{gehrig2018esim} & Mono & ESIM & SLAM & $752 \times 480$ & \xmark & \xmark \\
\textit{v2e}-Cityscapes~\cite{hu2021v2e} & Cityscapes~\cite{cordts2016cityscapes} & \textit{v2e} & Segmentation & $2048 \times 1024$ & \xmark & \xmark \\
\textit{v2e}-COCO~\cite{hu2021v2e} & COCO~\cite{lin2014microsoft} & \textit{v2e} & Object Detection & Varied & \xmark & \xmark \\ \midrule
\textbf{eSkiTB (Ours)} & \textbf{SkiTB}~\cite{Bachmann2019SkiPose} & \textbf{\textit{v2e}} & \textbf{Tracking} & \textbf{$1280 \times 720$} & \cmark & \cmark \\ \bottomrule
\end{tabular}%
}
\label{tab:simulated_datasets}
\end{table*}

\section{Related Study}
\label{sec:background}

\subsection{Frame-based Ski Tracking}
% Traditional winter sports datasets are predominantly confined to the RGB domain. Seminal works such as the SkiTB benchmark~\cite{Bachmann2019SkiPose} and recent large-scale collections like Youtube-SkiJump~\cite{DUNNHOFER2024103978} rely on standard frame-based cameras. While these datasets have advanced pose estimation and action recognition, they are fundamentally limited by the physics of the sensor.
% In the context of skiing, where athletes exceed $100$\, km/h against a complex background~\cite{Bachmann2019SkiPose}, standard sensors suffer from two critical failure modes: \textit{occlusion} by static broadcast elements, and \textit{distractor confusion} caused by background clutter. As noted in broader human activity datasets like \textit{NTU RGB+D}~\cite{Shahroudy2016NTURGBD}, scaling these datasets requires laborious manual annotation, which becomes infeasible when broadcast overlays obscure the ground truth.

Traditional winter sports datasets are predominantly RGB-based. Seminal works such as \textit{SkiTB}~\cite{Bachmann2019SkiPose} and recent large-scale collections like \textit{YouTube-SkiJump}~\cite{DUNNHOFER2024103978} rely on conventional frame cameras for pose estimation and action analysis. However, the physics of frame-based sensing create inherent limitations for high-speed skiing. Athletes frequently exceed $100$\, km/h~\cite{Bachmann2019SkiPose}, and broadcast footage introduces \emph{screen-space artifacts} (e.g., overlays) and \emph{world-space clutter} (e.g., banners), both of which obscure the athlete and degrade appearance-based tracking. As observed in other large-scale human activity datasets such as NTU RGB+D~\cite{Shahroudy2016NTURGBD}, annotation quality degrades sharply when visual distractors interfere with the target, making robust tracking and scalable labeling difficult. Table~\ref{tab:dataset_comparison} highlights how \textit{eSkiTB} fills the gap left by the traditional benchmarks.

\begin{figure}[t]
  \centering
  \includegraphics[width=0.9\linewidth]{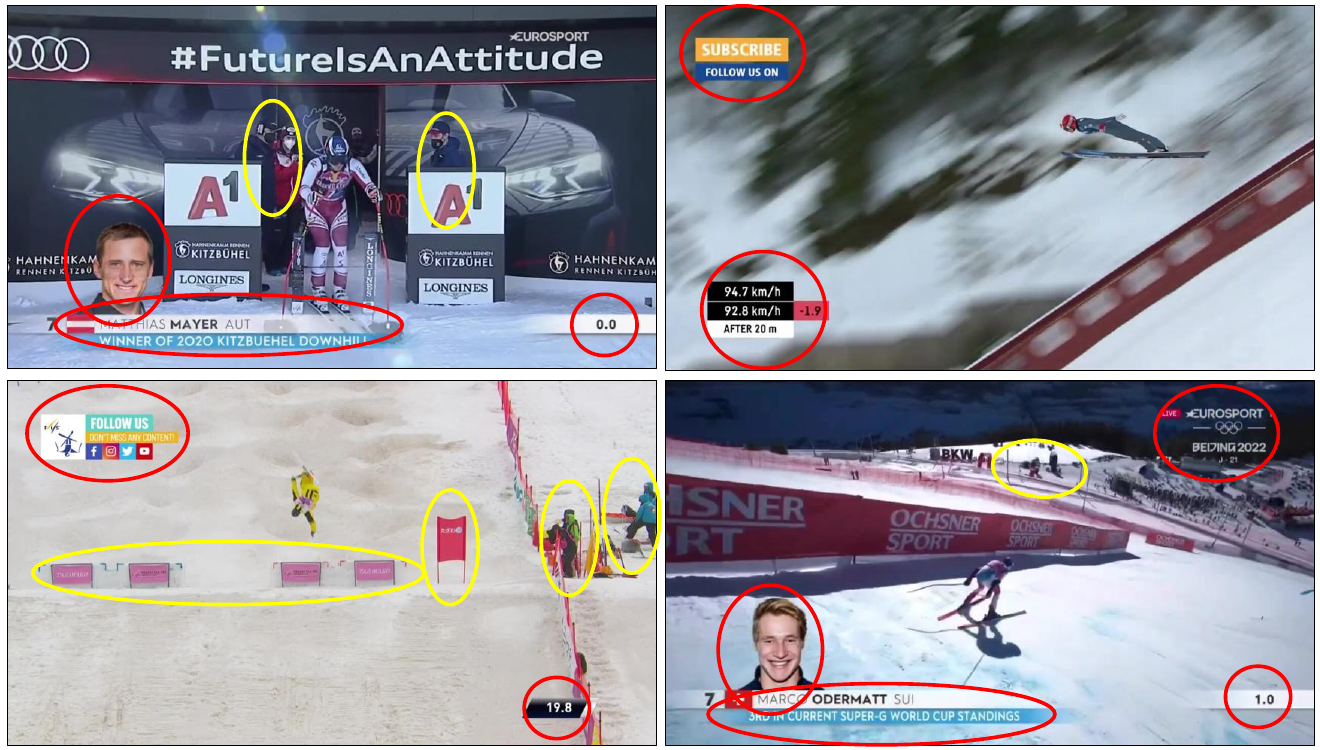}
  \caption{ Illustrates the weakness in  standard RGB images which suffers from \textit{screen-space artifacts} (yellow) and \textit{world-space clutter} (red). While these artifacts obscure the athlete in the intensity domain, our proposed event-based pipeline naturally filters the static screen-space overlays.}
  \label{fig:rgb_challenges}
\end{figure}

\subsection{Event-based Vision for Skiing}
Inspired by the successful adoption of event-based vision across multiple domains \cite{verma2024etram, aliminati2024sevd, chakravarthi2023event, chanda2025event, vinod2025sebvs, chanda2025sepose, tan2025real}, we explore its applicability to skier tracking.
Event cameras offer an alternative sensing modality that asynchronously records brightness~\cite{Gallego2020Survey, chakravarthi2024recent}, making them intrinsically robust to visual clutter common in skiing broadcasts. Two categories of distractors behave differently in the event domain:

\begin{enumerate}
    \item \textbf{\textit{Screen-space artifacts}}: Digital overlays that are fixed in image coordinates generate no temporal change, and thus, produce \emph{zero events} disappearing entirely from the event stream as shown in \autoref{fig:rgb_challenges}.
    \item \textbf{\textit{World-space clutter}}: Physical elements such as fences or banners generate events only during camera motion and therefore follow the global background flow.
\end{enumerate}

The skier, by contrast, exhibits \textit{independent motion}, producing a strong temporal-contrast signature that event-based trackers can isolate more reliably than RGB-based methods relying on texture.
% To generate synthetic events from broadcast footage, we evaluated ESIM~\cite{gehrig2018esim}, \textit{v2e}~\cite{hu2021v2e}, and the Prophesee simulator~\cite{finateu20205}. Consistent with prior findings as shown in Table~\ref{tab:simulated_datasets}, \textit{v2e} produced cleaner event contours and more stable temporal structure, particularly in the low-texture regions such as snow, as shown in Figure~\ref{fig:sim_comparison}. To standardize qualitative comparison across simulators, all event streams were visualized using Prophesee's Metavision toolkit~\cite{finateu20205}. A qualitative comparison of the frames generated from the simulated event streams from each simulator is show in \autoref{fig:sim_comparison}

\begin{figure}[t]
  \centering

   \includegraphics[width=0.9\linewidth]{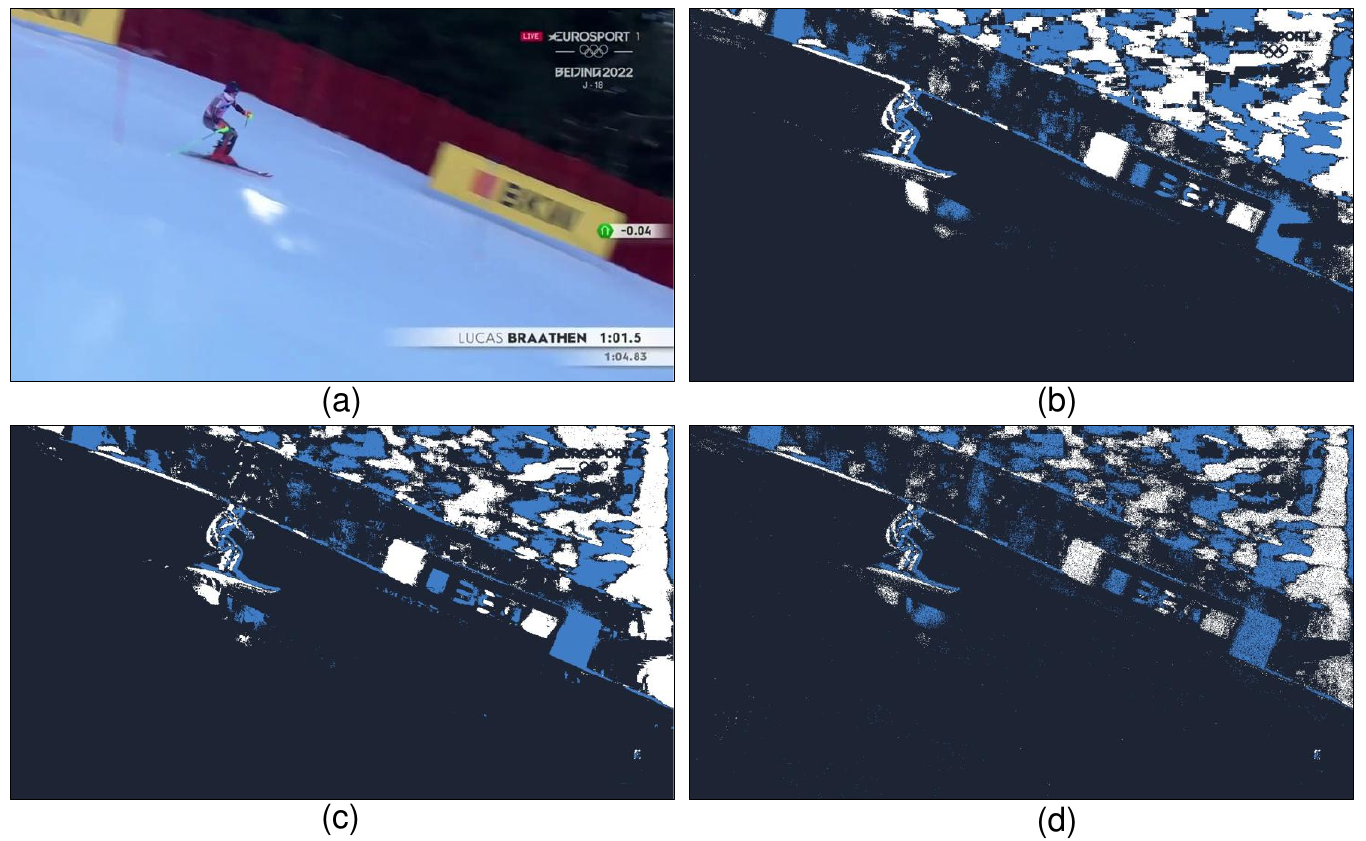}
   \caption{Qualitative comparison on (a) sample RGB ski-jump, (b) \textit{v2e} preserves the skier's silhouette and motion details better than (c) ESIM and (d) Prophesee's simulator, which exhibits significant noise and artifacts.}
   \label{fig:sim_comparison}
\end{figure}

To generate synthetic events from broadcast footage, each of the \textit{ESIM}~\cite{gehrig2018esim}, \textit{v2e}~\cite{hu2021v2e}, and the Prophesee ~\cite{finateu20205} simulators were utilized, without further frame based interpolation prior to generating the event streams. Prophesee's Metavision toolkit~\cite{finateu20205} was utilized to visualize and qualitatively compare the simulated event streams, as shown in \autoref{fig:sim_comparison}. Consistent with prior findings as shown in Table~\ref{tab:simulated_datasets}, \textit{v2e} produced cleaner event contours and more stable temporal structure, particularly in the low-texture regions such as snow.

\subsection{Ski Tracking Paradigms}

% To analyze the benefits of event data, we evaluate two state-of-the-art tracking architectures:

% \begin{itemize}
%     \item \textbf{STARK}~\cite{yan2021learning}: 
%     A spatio-temporal Transformer that represents the state of the art in RGB-based tracking. STARK relies on appearance cues, spatial attention, and online template updates, performing well in textured scenes but degrading in low-texture or cluttered environments.

%     \item \textbf{SDTrack}~\cite{shan2025sdtrack}:
%     A spiking Transformer designed for asynchronous event streams. SDTrack incorporates a global trajectory prompt (GTP) to maintain object permanence and exploits temporal contrast and sparsity, making it inherently more robust to clutter and background motion.
% \end{itemize}

To study the impact of sensing modality, we benchmark two complementary tracking architectures:

\begin{itemize}
    \item \textbf{\textit{STARK}}~\cite{yan2021learning}: A SOTA RGB Transformer relying on appearance cues, spatial attention, and template updates. \textit{STARK} performs strongly in textured scenes but degrades under clutter and low-texture conditions typical of winter sports.
    \item \textbf{\textit{SDTrack}}~\cite{shan2025sdtrack}: A Spiking Transformer tailored for asynchronous event streams. \textit{SDTrack} employs a global trajectory prompt (GTP) to maintain object permanence and exploits temporal contrast, offering inherent robustness to distractors and ego-motion.
\end{itemize}

\begin{figure*}[t]
  \centering
  \includegraphics[width=0.98\linewidth]{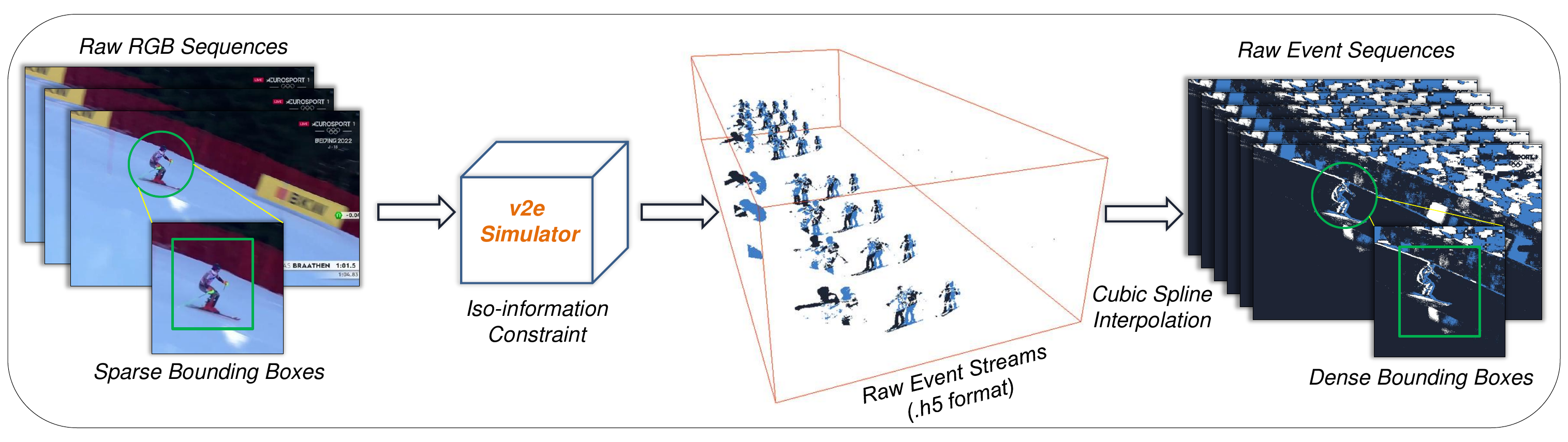}
   \caption{Illustrates \textit{eSkiTB} generation pipeline. We process raw high-resolution RGB frames through the \textit{v2e} simulator to generate asynchronous event streams. Crucially, we bypass neural frame interpolation to maintain an iso-informational constraint, ensuring that the event data contains no hallucinated temporal information. Ground truth bounding boxes are interpolated to $1$\, ms resolution.}
   \label{fig:pipeline}
\end{figure*}

\subsection{SkiTB as the Foundation}
The SkiTB benchmark~\cite{Bachmann2019SkiPose} serves as the ideal foundation for our event-based conversion. Unlike general-purpose datasets, SkiTB provides high-resolution, high frame rate footage specifically curated for ski jumping analysis. Its sequences capture the full ballistic trajectory of the athlete, from the in-run to the landing, against the challenging backdrop of snow and broadcast overlays. Using these specialized RGB data, we can generate a synthetic event dataset that rigorously tests the capabilities of event-based trackers in a domain where they are theoretically superior.

\section{The \textit{eSkiTB} Dataset}
\label{sec:dataset}

The primary contribution of this work is the formulation of \textit{eSkiTB}, a neuromorphic benchmark generated under strict signal-processing constraints. The pipeline transforms standard broadcast footage into high-fidelity event streams without introducing synthetic temporal artifacts.

\subsection{Source Data and Diversity}
\label{subsec:source_diversity}
The \textit{eSkiTB} dataset is derived from the SkiTB benchmark~\cite{skitb_wacv2024}, comprising $300$ high-resolution video sequences of competitive skiing across three disciplines: alpine skiing (AL), freestyle skiing (FS), and ski jumping (JP). A key strength of the dataset is its environmental diversity, featuring recordings from $98$ unique geographical locations worldwide. The footage captures a wide spectrum of atmospheric conditions, classified into $10$ distinct weather categories ranging from clear sunny skies (high glare) to heavy snowfall, fog, and rain (low contrast). Furthermore, the dataset includes both day and night events, introducing significant variations in lighting that are particularly relevant for event-based vision sensors.

\subsection{Event Simulation Under the Iso-Informational Constraint}
\label{subsec:iso_informational}
A critical methodological deviation in our pipeline is the strict rejection of neural frame interpolation. Prior video-to-event (\textit{v2e}) works often employ networks like SuperSloMo~\cite{jiang2018super} to upsample $30$\,\textit{fps} video to $>1000$\,\textit{fps} before conversion. While this yields smoother visualizations, it introduces synthetic temporal data that does not exist in the source signal.
To ensure a rigorous scientific control, we enforce an \textit{iso-informational constraint}, strictly avoiding neural upsampling to ensure \textit{photometric purity}. We feed the raw, temporally quantized RGB frames directly into the \textit{v2e} simulator~\cite{hu2021v2e}. While this quantizes the motion to the source video framerate, it guarantees that every event in the dataset corresponds to a real, observed photometric change, rather than a predicted one. This establishes a \textit{conservative lower bound} for tracker performance; algorithms that succeed on this quantized signal are expected to perform even better on native high-speed hardware.

\begin{figure*}[t]
  \centering
  \includegraphics[width=0.95\linewidth]{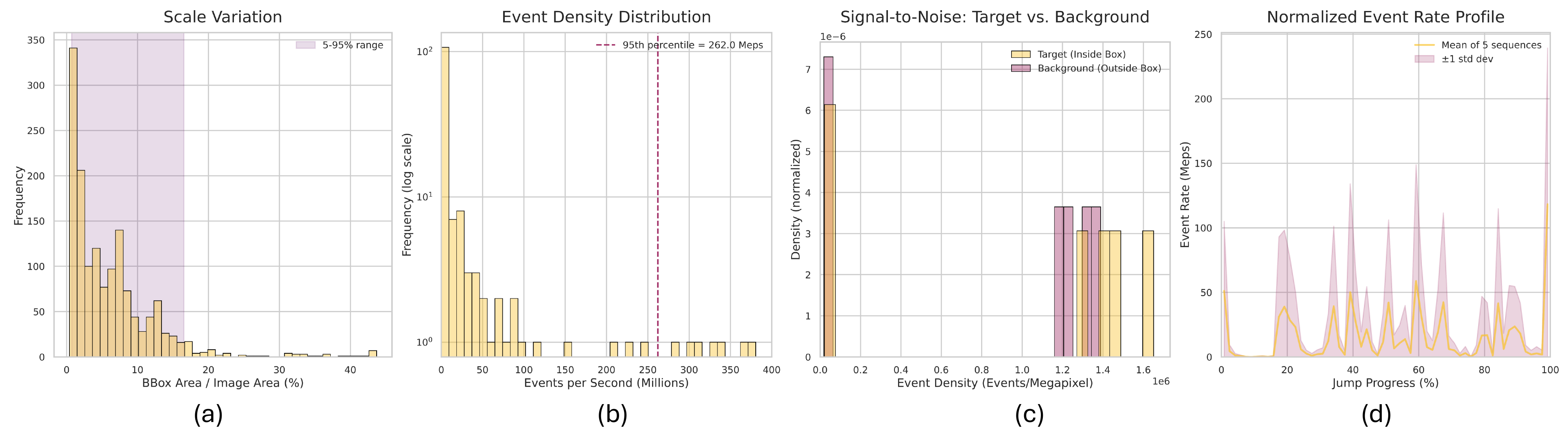}
   \caption{Represents \textit{eSkiTB} Statistics. (a) Density: Heavy-tailed global rates (95th percentile: $1.58$, Meps) confirm high-dynamic motion. (b) SNR: Consistent shift to higher target densities proves natural segmentation from background clutter. (c) Scale: Skewed bbox area ($< 4\%$ of image) reflects extreme zoom-induced variance. (d) Temporal: Information density peaks during critical take-off ($\approx 20\%$) and flight.}
   \label{fig:stats}
\end{figure*}

\subsection{Data Generation Pipeline}
We utilize the \textit{v2e} pipeline on raw frames without intermediate synthesis as shown in \autoref{fig:pipeline}
~\cite{hu2021v2e, vinod2025sebvs}.
\begin{itemize}
    \item \textbf{\textit{Configuration:}} We utilize the photoreceptor exposure mode to accurately model pixel intensity changes over the exposure duration. The simulation is configured with a standard contrast threshold of $C_{pos} = C_{neg} = 0.2$. To ensure realistic noise modeling, we configured the simulator with a leak rate of $\tau_{leak}=0.01$\, Hz and a photoreceptor cutoff frequency of $f_{3dB}=300$\, Hz. Photon shot noise was enabled ($0.001$\, Hz) to mimic the challenging low-light conditions often found in overcast winter broadcasts.
    \item \textbf{\textit{Reproducibility:}} By disabling slomo interpolation, the conversion becomes deterministic and strictly faithful to the original visual signal.
\end{itemize}

\subsection{Data Composition and Structure}
\label{subsec:data_structure}
The \textit{eSkiTB} dataset comprises a total of $300$ sequences, totaling approximately $235$ minutes of high-speed footage, partitioned into:
\begin{itemize}
    \item \textbf{\textit{Train split:}} $240$ sequences, used for training the SNN backbone.
    \item \textbf{\textit{Validation split:}} $30$ sequences, used for hyperparameter tuning and checkpoint selection.
    \item \textbf{\textit{Test split:}} $30$ sequences, held out strictly for final evaluation.
\end{itemize}

The dataset is organized to facilitate efficient streaming and random access:
\begin{itemize}
    \item \textbf{\textit{Event streams:}} Raw events are stored in HDF5 format at the full source resolution of $1280 \times 720$. We release the dataset at this high definition to future-proof the benchmark, enabling next-generation trackers to exploit fine-grained spatial details. Each file contains the event tuples $(t, x, y, p)$, where $t$ is the timestamp in microseconds, $(x, y)$ are the pixel coordinates, and $p \in \{0, 1\}$ represents the polarity (OFF/ON).
    \item \textbf{\textit{Annotations:}} Ground truth bounding boxes are provided for the single class \textit{`Skier'}, defined in JSON format. We supply both event-aligned annotations (aligned to the original video frames) and dense interpolated annotations (interpolated at $1$\, ms intervals). The bounding boxes are defined in absolute pixel coordinates $[x, y, w, h]$ relative to the $1280 \times 720$ resolution, where $(x, y)$ denotes the top-left corner.
\end{itemize}

\subsection{Dataset Statistics}
\begin{figure*}[t!]
  \centering
   \includegraphics[width=0.85\linewidth]{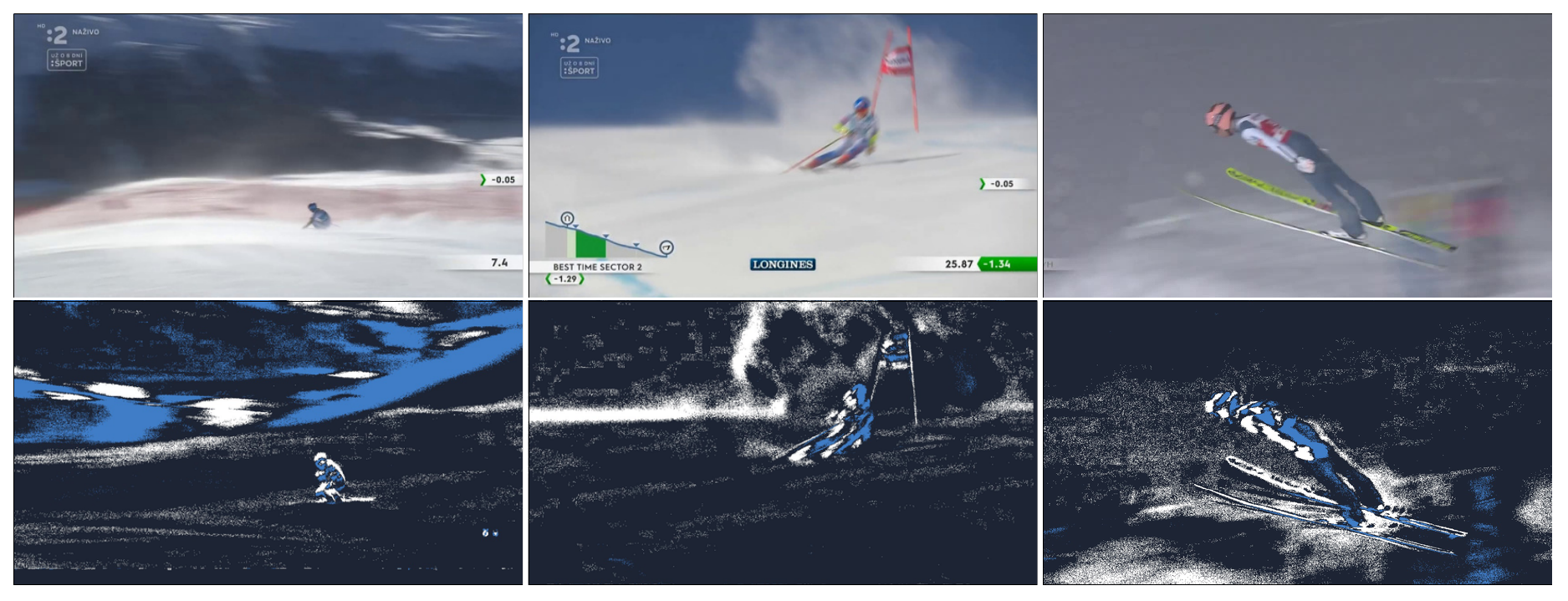}
   \caption{Qualitative samples from the \textit{eSkiTB} dataset. Top row: Original RGB frames showing the challenging conditions with background clutter and broadcast overlays. Bottom row: Corresponding event representations accumulated over $30$\, ms. The event stream successfully captures the motion of the skier (high contrast) while differentiating it from background elements.}
   \label{fig:qualitative_samples}
\end{figure*}

To characterize the complexity of \textit{eSkiTB}, we analyze the event density, motion profiles, and target variations. The dataset features high event density in regions of skier motion, contrasting sharply with the sparse background events generated by the snow, as shown in \autoref{fig:qualitative_samples}. The average event rate exceeds $10 \times 10^6$ events per second, with peaks significantly higher during the take-off and 
landing phases. \textit{eSkiTB} presents unique challenges due to extreme variability.
\begin{itemize}
    \item \textbf{\textit{Scale variation:}} The dataset presents a significant multi-scale tracking challenge. Due to broadcast camera zooming, the target bounding box area varies by orders of magnitude, ranging from $0.001\%$ of the image area at the take-off gate to $100\%$ during extreme close-ups, with a mean coverage of $4.99\%$.

    \item \textbf{\textit{Long-term tracking:}} Unlike typical short-term tracking benchmarks, \textit{eSkiTB} focuses on endurance. Sequence lengths range from $275$ to $3,582$ frames, with an average duration of $1,176$ frames. These extended-duration tests the tracker's ability to maintain lock over long periods without drifting.
\end{itemize}

\subsection{Dense Temporal Ground Truth}
Standard computer vision datasets \cite{geiger2013vision,skitb_wacv2024} provide annotations at the frame rate ($30$--$60$\,Hz). However, event-based trackers operate in the microsecond regime as shown in \autoref{fig:stats}. To bridge this temporal gap, we define dense ground-truth labels at $1$\, ms intervals ($1000$\, Hz) via interpolation.

We provide ground truth bounding boxes aligned with the original video frames. For evaluation, let $(\mathbf{b}_i, t_i)$ and $(\mathbf{b}_{i+1}, t_{i+1})$ denote the bounding box coordinates and timestamps of two consecutive labeled frames. For any intermediate timestamp $t \in [t_i, t_{i+1}]$, we compute the dense label $\mathbf{b}(t)$ via \textit{cubic spline interpolation} to approximate the ballistic trajectory. Unlike linear interpolation, which introduces unrealistic \textit{``zigzag''} artifacts at keyframes, cubic splines model the smooth, continuous nature of ballistic flight. This is physically mandated by the aerodynamic forces (drag and lift) acting on the ski jumper, ensuring that the ground truth trajectory respects the $C^2$ continuity of the athlete's center of mass.

\begin{equation}
    \mathbf{b}(t) = \text{Spline}(t; \{\mathbf{b}_k, t_k\})
    \label{eq:interpolation}
\end{equation}

This continuous-time supervision allows us to calculate Intersection over Union (IoU) at the native temporal resolution of the event sensor, ensuring that trackers are penalized for drift that occurs \textit{between} RGB frames.

\subsection{Quality and Fidelity Diagnostics}
To verify the integrity of the conversion, we perform a series of diagnostic checks. We compute the cumulative event count per pixel over the entire sequence duration. This \textit{``event image"} must reconstruct the visual structure of the scene (e.g., the ski jump ramp, the athlete's trajectory) solely from temporal contrast changes. If the event count is uniform or sparse in high-motion regions, the sequence is flagged for manual inspection. Additionally, we verify that the event timestamps are strictly monotonic and aligned with the interpolated bounding box timestamps.

\section{\textit{eSkiTB} Evaluation}
\label{sec:evaluation}

To empirically validate the hypothesis that neuromorphic vision offers superior robustness in cluttered, high-speed environments, we conduct a comparative benchmark between SOTA frame-based and event-based trackers.

\subsection{Implementation Details}
\label{subsec:implementation}

To isolate the impact of sensory modality from architectural variance, we enforce \textit{Architectural Parity} by selecting two Transformer-based trackers: \textit{STARK}~\cite{yan2021learning} (RGB) and \textit{SDTrack}~\cite{shan2025sdtrack} (Event). Benchmarking a Spiking CNN against an RGB Transformer would conflate sensor benefits with backbone capacity. By comparing two Transformers, we ensure performance divergences are attributable to the input modality (temporal contrast vs. spatial appearance).

Our event-based tracking pipeline adapts the \textit{SDTrack} architecture. While \textit{eSkiTB} provides data at $1280 \times 720$, current SOTA SNNs are computationally limited to smaller input tensors. To reflect these hardware constraints, the continuous event stream is discretized into voxel grids ($128 \times 128 \times T$). We evaluate at this reduced resolution to benchmark current capabilities, leaving the utilization of the full high-definition stream for future work.

We fine-tune the \textit{SDTrack} base model (pretrained on generic event datasets) on the \textit{eSkiTB} training split. The training objective combines GIoU, L$1$, and location losses. We utilize the AdamW optimizer~\cite{loshchilov2017decoupled} with a learning rate of $1 \times 10^{-4}$ and a weight decay of $1 \times 10^{-4}$. The batch size is set to $2$ to maintain training stability given the computational constraints of the spatiotemporal attention mechanism; all experiments are verified to be conducted on a single \textit{NVIDIA RTX 4090 GPU}.
To mitigate overfitting, a prevalent challenge when adapting high-capacity transformers to specialized domains, we monitored validation loss throughout the training schedule. We observed that the model's generalization capability peaked at epoch $20$, after which the training loss continued to decrease, while the validation performance plateaued. Consequently, we utilize the checkpoint from epoch $20$ for all reported evaluations.

\subsection{Quantitative Results}
We adhere to the one-pass evaluation (OPE) protocol established by \textit{SkiTB}: trackers are initialized with the first ground-truth box of every multi-camera sequence, executed once without resets, and scored by the per-sequence averages of mean IoU, Precision@$20$\,px, and Success@$0.5$ IoU. The evaluation is performed on the official \textit{eSkiTB} test split, comprising $30$ multi-camera sequences across all disciplines.

\textbf{\textit{The necessity of domain adaptation.}} To quantify the domain gap between generic event vision and the adversarial winter sports environment, we first evaluated the off-the-shelf \textit{SDTrack} model without any fine-tuning. As shown in Table~\ref{tab:tracking_results}, this pre-trained baseline achieves a Mean IoU of only $0.312$. This poor performance confirms that generic event representations are insufficient for the \textit{``White Room"} effect, validating the necessity of the \textit{eSkiTB} dataset for domain-specific learning.
Table~\ref{tab:tracking_results} presents the comprehensive comparison. Despite \textit{STARK$_{\text{ft}}$} and \textit{STARK$_{\text{ski}}$} benefiting from RGB fine-tuning in the target domain, our fine-tuned neuromorphic pipeline significantly narrows the performance gap with skier-specific trackers and outperforms the standard \textit{STARK} baseline by $+19.9$ IoU points.

\begin{table}[t]
\centering
\resizebox{1\linewidth}{!}{%
\begin{tabular}{@{}lcccc@{}}
\toprule
\textbf{Tracker} & \textbf{Modality} & \textbf{Mean IoU} & \textbf{Precision@20px} & \textbf{Success@0.5} \\ \midrule
STARK (Generic)~\cite{yan2021learning} & RGB & $0.512$ & $0.567$ & $0.568$ \\
STARK (Fine-tuned)~\cite{yan2021learning} & RGB & $0.795$ & $0.847$ & $0.904$ \\
STARK (Ski-Specific)~\cite{yan2021learning} & RGB & \textbf{0.829} & \textbf{0.887} & \textbf{0.935} \\ \midrule
SDTrack (Pretrained) & Event & $0.312$ & $0.354$ & $0.418$ \\
\textbf{SDTrack (Finetuned)} & \textbf{Event} & $0.711$ & $0.720$ & $0.873$ \\
\bottomrule
\end{tabular}%
}
\caption{One-pass evaluation on the \textit{eSkiTB} test split. We include the pre-trained \textit{SDTrack} baseline to highlight the significant performance gain ($+0.399$ IoU) achieved via fine-tuning on our dataset. RGB trackers are reproduced from the publicly released \textit{SkiTB} results.}
\label{tab:tracking_results}
\end{table}

\textbf{\textit{Discipline-specific performance.}} \textit{SDTrack} demonstrates exceptional performance on JP, achieving $0.974$ IoU, $0.982$ Precision@$20$\,px, and $0.998$ Success. This represents a $+5.9$ IoU improvement over \textit{STARK$_{\text{ski}}$} and a substantial $+35.6$ point gain over the generic \textit{STARK} baseline. The model remains competitive on AL sequences ($0.762$ IoU vs. $0.815$ for \textit{STARK$_{\text{ft}}$)}. However, performance degrades in FS scenarios ($0.569$ IoU vs. $0.728$ for \textit{STARK$_{\text{ft}}$)}. The fixed spatial resolution of the voxel grid ($128 \times 128$) acts as a low-pass spatial filter. In FS, where the target's angular velocity is highest and the scale is smallest (during high jumps), this quantization leads to aliasing, causing the tracker to lose lock. This confirms the need for multi-scale or adaptive event representations in future work. It is important to note that the dataset itself supports the full $1280 \times 720$ resolution; the $128 \times 128$ bottleneck is purely an architectural limitation of the current \textit{SDTrack} baseline.

\begin{figure}[t]
  \centering
  \includegraphics[width=1 \linewidth]{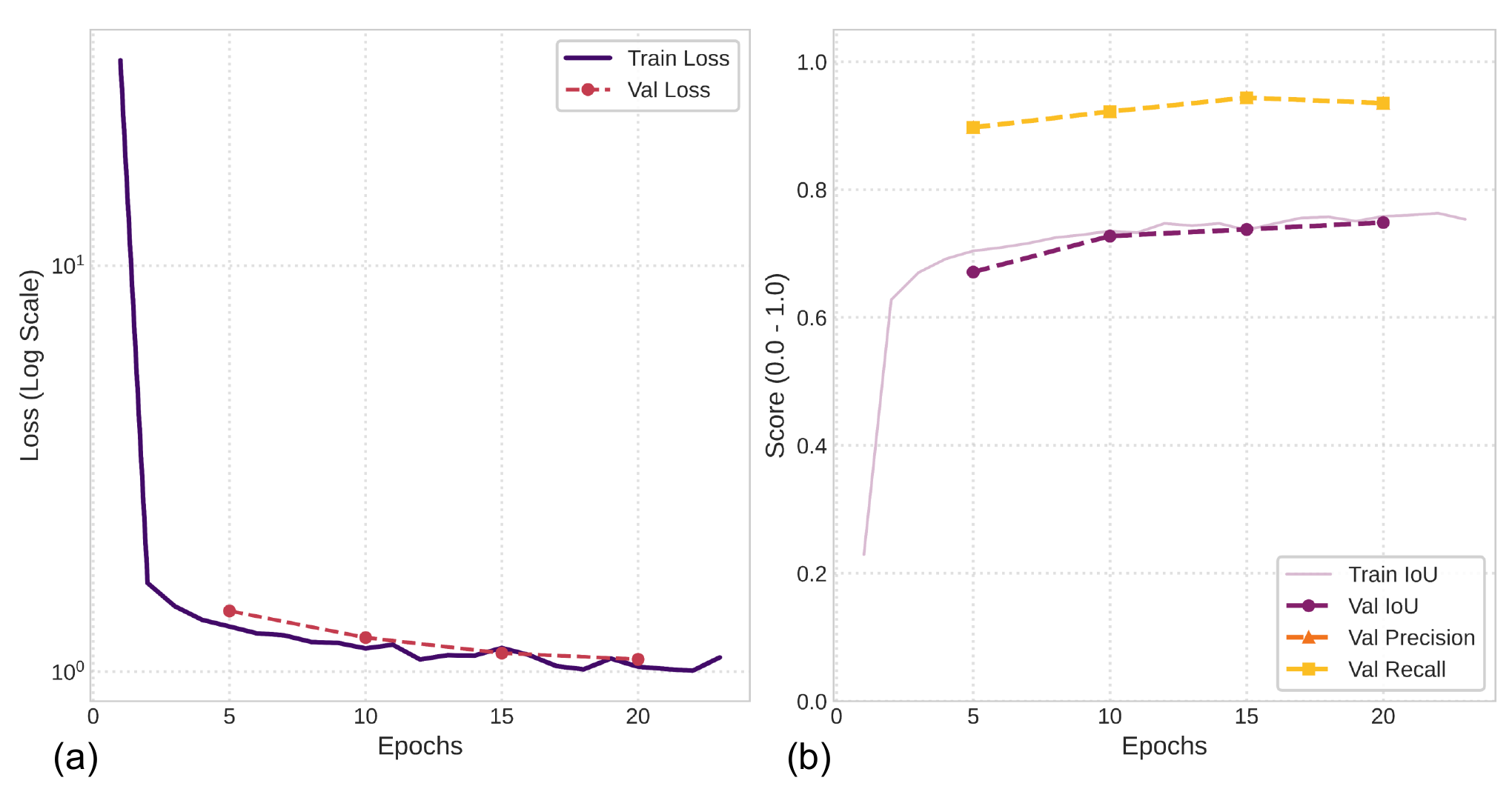}
   \caption{Training and validation dynamics of the \textit{SDTrack} model. 
(a) Training and validation losses (GIoU, L1, total) show stable convergence. 
(b) Validation metrics (success, precision, and normalized precision) peak around epoch 20, indicating the optimal checkpoint. 
(\textit{Validation precision and recall overlap in the plot.})
}
   \label{fig:performance_curves}
\end{figure}

\begin{figure*}[t]
  \centering
   \includegraphics[width=0.75\linewidth]{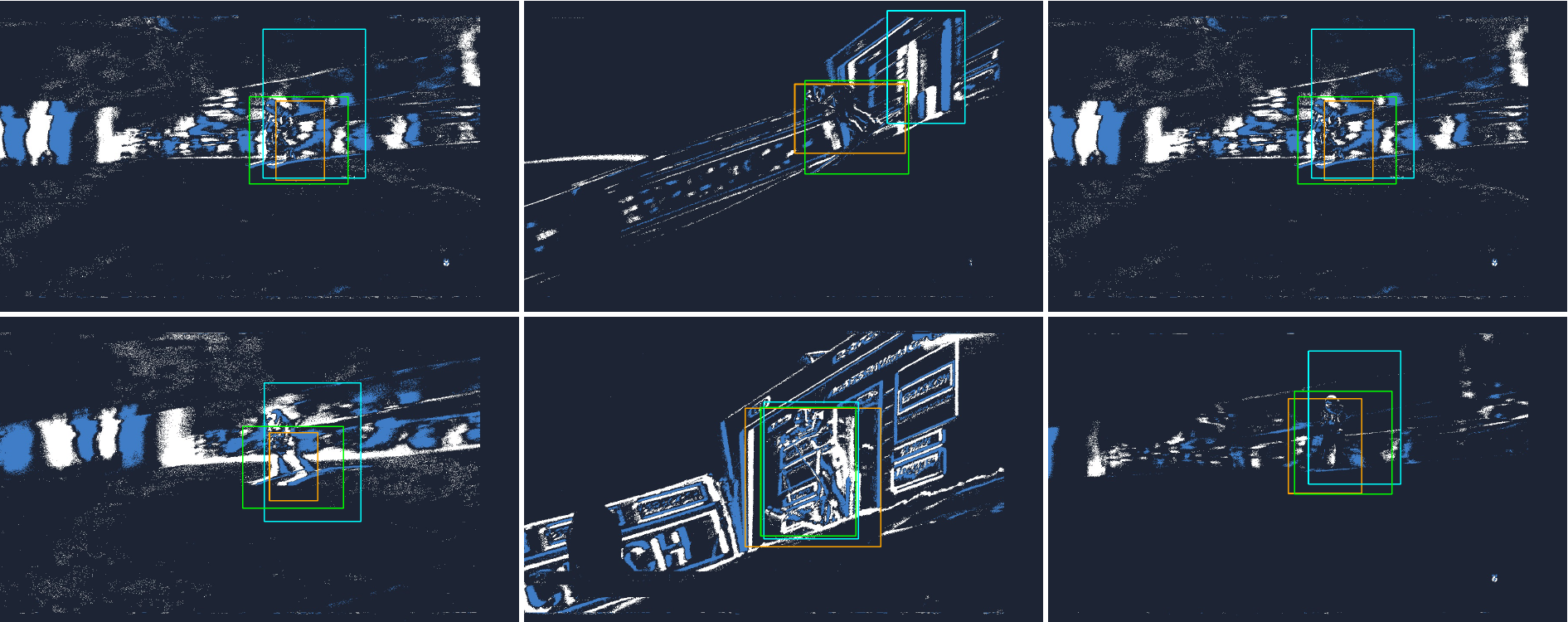}
   \caption{Qualitative performance of  \textit{SDTrack} model, with ground truth annotations shown in green and predictions in yellow.}
   \label{fig:qualitative_tracking}
\end{figure*}

\textbf{\textit{Attribute-conditioned analysis.}} By intersecting the test split with the attribute annotations provided by \textit{SkiTB}, we observe distinct modality-specific behaviors. Notably, in extreme \textit{low-resolution} sequences, \textit{SDTrack} maintains an IoU above $0.72$, whereas the RGB \textit{STARK} baseline degrades significantly to $0.47$--$0.59$ IoU. Conversely, sequences tagged with both \textit{low-resolution} and \textit{fast motion}, such as certain freestyle clips, remain challenging for both modalities, suggesting the need for an event-native aerial motion prior.

\subsection{The Clutter Stress Test}
To rigorously quantify robustness against visual distractors, we construct a \textit{``High-Clutter''} split by filtering for test sequences where single-camera clips exhibit at least $50\%$ \textit{background clutter} or \textit{partial occlusion} annotations. This subset comprises nine multi-camera sequences characterized by dense visual noise.
On this challenging subset, \textit{SDTrack} achieves $0.685$ IoU, $0.694$ precision@$20$\,px, and $0.860$ success@$0.5$. This corresponds to a $+20.0$ IoU, $+15.2$ precision, and $+33.8$ success point gain over the generic \textit{STARK} baseline. These results empirically demonstrate that temporal-contrast sensing is robust to visual distractors. While physical clutter (banners) generates high event density due to camera panning, the SNN effectively disentangles the \textit{Global Optical Flow} of the background from the \textit{Local Motion} of the athlete. Furthermore, the complete absence of events from static digital overlays removes a significant source of occlusion that plagues RGB trackers. While skier-optimized \textit{STARK} variants retain a performance advantage (e.g., $0.833$ IoU for \textit{STARK$_{\text{ski}}$}), the gap narrows to approximately $10$ IoU points, despite the RGB trackers having access to the full visual texture of the training set.

Figure~\ref{fig:performance_curves} illustrates the temporal stability of the trackers. In sequences with camera pans across high-contrast static elements, \textit{STARK} tends to drift towards sponsor banners. In contrast, \textit{SDTrack} maintains target lock by leveraging the motion-only signal. Figure~\ref{fig:qualitative_tracking} provides a qualitative comparison, highlighting the event-based tracker's robustness in clutter versus its limitations during complex aerial spins.

\section{Discussion}
\label{sec:discussion}

\begin{figure}[t]
  \centering
  \includegraphics[width=0.8\linewidth]{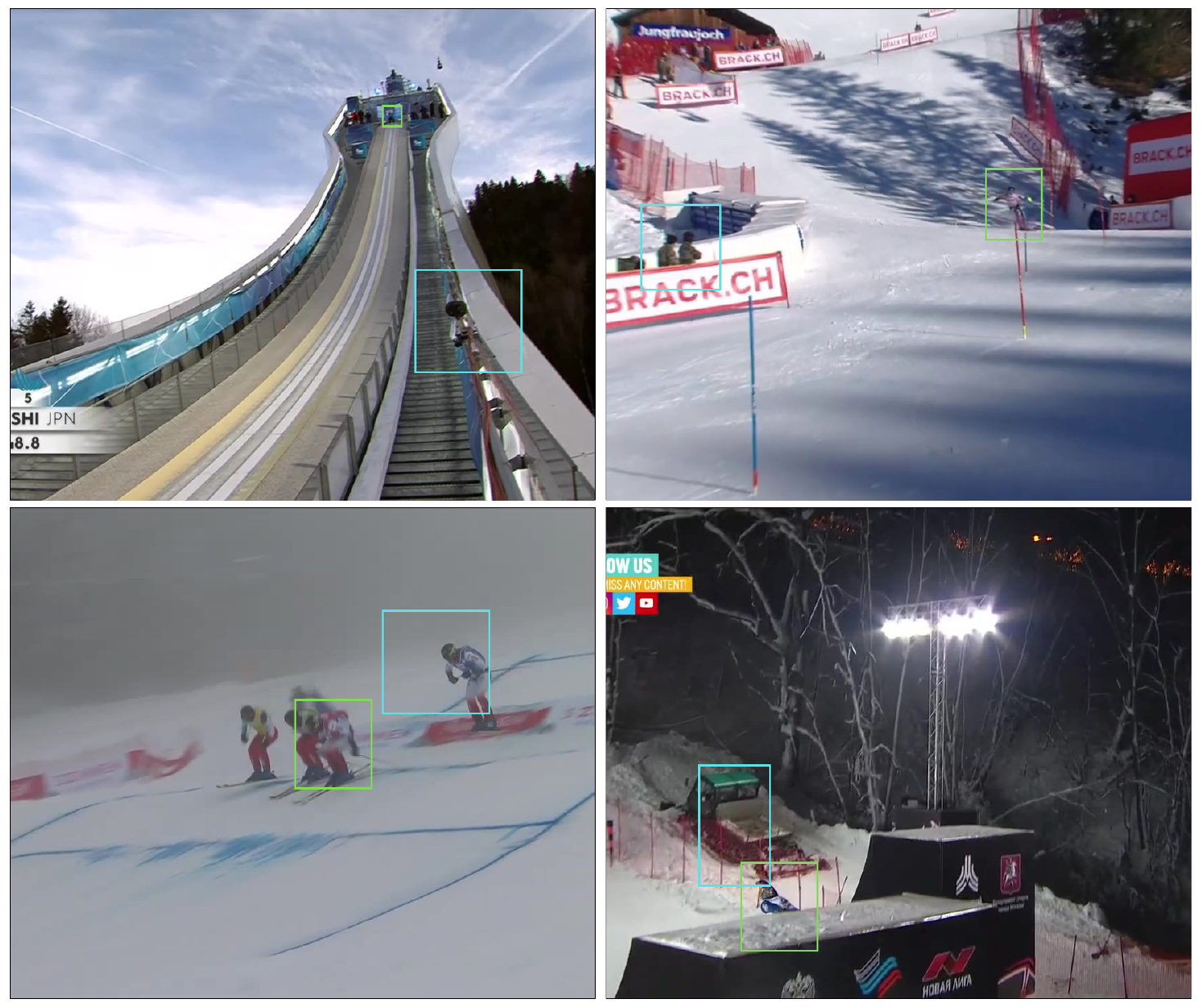}
  \caption{RGB trackers fail under broadcast clutter: static overlays and banners dominate appearance cues, causing the drift effect.}
  \label{fig:rgb_fails}
\end{figure}

% Our results show that in high-albedo winter environments, temporal contrast offers a distinct architectural advantage over spatial texture. By sensing motion directly, event cameras effectively solve the \textit{clutter disambiguation} problem, rendering trackers invariant to static banners that confuse RGB models. This has immediate implications for autonomous drone cinematography, suggesting that neuromorphic control loops could enable safer following in complex alpine terrain.
Our results show that in high-albedo winter environments, temporal contrast provides a fundamentally different and in many cases superior signal for tracking compared to spatial appearance. Event-based trackers benefit from directly sensing motion, enabling them to resolve the \textit{clutter disambiguation} challenge (see Figure \ref{fig:rgb_fails}) that hinders RGB models. Static broadcast elements such as overlays or banners generate no events and therefore vanish from the event stream, allowing the tracker to focus on the athlete’s independent motion. This observation has broader implications for applications such as autonomous ski-filming drones, where neuromorphic control loops could offer safer and more reliable following in visually congested alpine terrain.
A key limitation of \textit{eSkiTB} is the temporal resolution of the source RGB ($25$--$60$\,Hz). Since we disable neural interpolation to avoid hallucination, our simulated events represent a linearized approximation of the trajectory, smoothing out micro-dynamics ($<40$\,ms). Critically, this implies our benchmark serves as a \textit{conservative lower bound}. Real event cameras would capture these missing high-frequency signals, theoretically yielding even higher discriminative power than reported here. Therefore, the primary contribution of this work is not the microsecond temporal resolution itself, but the demonstration of event-based robustness to high dynamic range and background clutter, which remains valid even with the temporal quantization.
Furthermore, \textit{eSkiTB} is designed to stress-test event representations. The performance drop in \textit{freestyle skiing} exposes the fragility of fixed-grid voxelization against large aerial excursions. Rather than tuning hyperparameters to \textit{``solve''} the benchmark, we report this sensitivity to motivate research into adaptive event representations. Future work will focus on validating these synthetic results by collecting a smaller, real-world validation set. Due to the strict prohibition of non-broadcast drone flights during competitive events, simultaneous real-world event data collection remains a logistical challenge. Also, we deliberately exclude hybrid baselines to isolate the \textit{marginal utility} of the asynchronous modality. In resource-constrained applications like drones, the goal is often to \textit{offload or augment} heavy RGB processing. Our experiments demonstrate that in high-clutter regimes, event-only tracking outperforms RGB-only tracking without the bandwidth cost of sensor fusion.
% \section{Conclusion}
% \label{sec:conclusion}

% This work addresses the fundamental computer vision bottleneck in professional winter sports: the inability of standard frame-based sensors to robustly track high-speed motion in cluttered, high-speed broadcast environments. We introduced \textit{eSkiTB} , the first large-scale neuromorphic benchmark tailored for this adversarial domain.

% By strictly enforcing an \textit{Iso-Informational Constraint} during the generation pipeline, we established a rigorous experimental control that precluded synthetic data hallucination. This allowed for a fair, ``modality-vs-modality'' assessment. Our experiments demonstrated that while state-of-the-art RGB trackers (STARK) suffer from catastrophic drift when the athlete is lost in background clutter and broadcast overlays, Spiking Transformers (SDTrack) maintain robust lock by exploiting the distinct spatiotemporal signature of the athlete's independent motion against the global background flow.

% We conclude that the future of autonomous sports analytics, particularly for drone-based cinematography in alpine environments, lies in the asynchronous domain. \textit{eSkiTB} serves as the foundational testbed to accelerate this transition. To facilitate further research, we will release the complete dataset, the Iso-Informational conversion pipeline, and the SNN evaluation checkpoints under an open-source license.

\section{Conclusion}
\label{sec:conclusion}

This work tackles a core limitation in winter-sports vision: the inability of RGB-based trackers to robustly localize high-speed athletes in cluttered broadcast footage. We introduced \textit{eSkiTB}, the first neuromorphic benchmark tailored to this domain, and generated its event streams under a strict \textit{iso-informational constraint} to ensure a fair comparison between sensing modalities. Our experiments show that while SOTA RGB trackers such as \textit{STARK} fail under broadcast overlays and texture collapse in RGB, the Spiking Transformer \textit{SDTrack} maintains stable tracking by exploiting the athlete’s distinct temporal-contrast signature while tested on events. These findings demonstrate the potential of asynchronous sensing for future autonomous sports analytics and drone-based cinematography in alpine environments. By releasing the dataset, conversion pipeline, and evaluation code. Thus serving as the foundation of event-based vision in skiing.
{
    \small
    \bibliographystyle{ieeenat_fullname}
    \bibliography{main}
}

\end{document}